# Multi-Branch Fully Convolutional Network for Face Detection

Yancheng Bai
KAUST
yancheng.bai@kaust.edu.sa

Bernard Ghanem
KAUST
bernard.ghanem@kaust.edu.sa

## Abstract

*Face detection is a fundamental problem in computer vision. It is still a challenging task in unconstrained conditions due to significant variations in scale, pose, expressions, and occlusion. In this paper, we propose a multi-branch fully convolutional network (MB-FCN) for face detection, which considers both efficiency and effectiveness in the design process. Our MB-FCN detector can deal with faces at all scale ranges with only a single pass through the backbone network. As such, our MB-FCN model saves computation and thus is more efficient, compared to previous methods that make multiple passes. For each branch, the specific skip connections of the convolutional feature maps at different layers are exploited to represent faces in specific scale ranges. Specifically, small faces can be represented with both shallow fine-grained and deep powerful coarse features. With this representation, superior improvement in performance is registered for the task of detecting small faces. We test our MB-FCN detector on two public face detection benchmarks, including FDDB and WIDER FACE. Extensive experiments show that our detector outperforms state-of-the-art methods on all these datasets in general and by a substantial margin on the most challenging among them (e.g. WIDER FACE Hard subset). Also, MB-FCN runs at 15 FPS on a GPU for images of size $640 \times 480$ with no assumption on the minimum detectable face size.*

## 1. Introduction

Face detection is a fundamental and important problem in computer vision, since it is usually a key step towards many subsequent face-related applications, including face parsing, face verification, face tagging and retrieval, etc. Face detection has been widely studied over the past few decades and numerous accurate and efficient methods have been proposed for mostly constrained scenarios. Recent works focus on faces in uncontrolled settings, which is much more challenging due to the significant variations in illumination, pose, scale and expressions. A thorough survey on face detection methods can be found in [35].

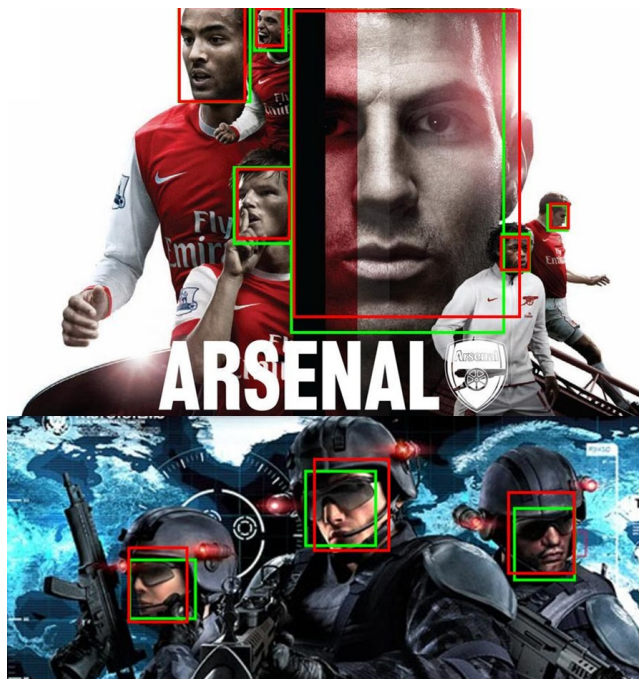

Figure 1. Example detection results generated by our proposed MB-FCN face detector. It can handle faces with large scale variations, extreme pose, exaggerated expressions, and occlusion with only a single pass through the backbone network. Bounding boxes with a green and red color are ground-truth annotations and MB-FCN detection results, respectively.

Face detection is the task of finding the locations of all faces in an image with arbitrary sizes. As shown in Figure 1, the scale variations may be significant. To deal with this problem, both traditional face detectors [27, 4, 29] and CNN-based ones [13, 28, 40] have to exhaustively search for faces on different levels of an image pyramid constructed from the original image. This constitutes the main computational bottleneck of many modern face detectors, which prohibit their use in a vast range of realtime real-world applications. Moreover, small faces tend to require higher-resolution features for discrimination and effective localization. However, most CNN-based detectors represent targets by only exploiting the deep coarse features, which lose most spatial information of small faces due to the down-sampling operations. This prohibits accurate localization of these types of challenging faces, which is reflected in the substantial state-of-the-art performance disparity between the detection of small vs. large faces.

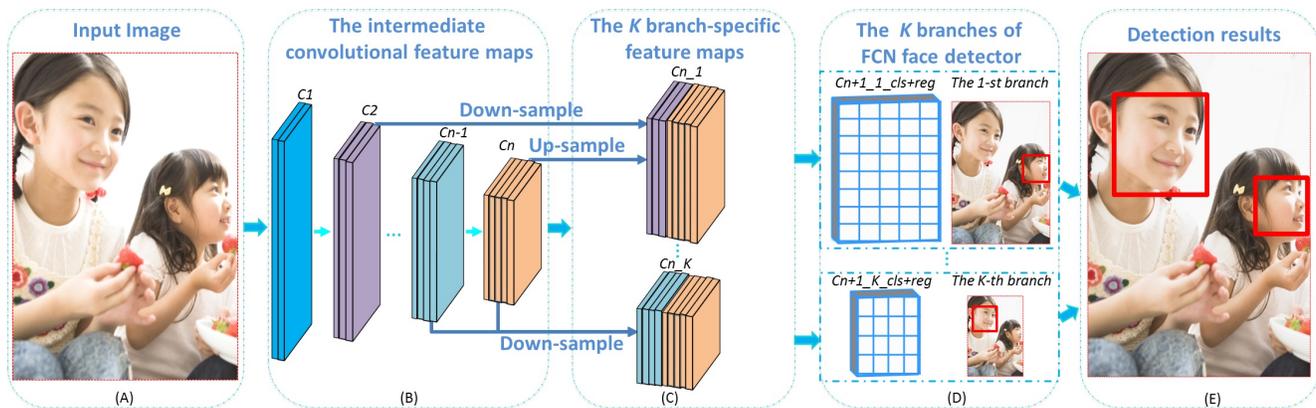

Figure 2. The pipeline of the proposed multi-branch FCN face detector system. (A) Testing images are fed to the network; (B) After several layers of convolution, a series of *conv* feature maps at several scales are generated; (C) Scale specific skip connections of *conv* feature maps generate $K$ specific feature maps for each branch to represent faces at specific scale ranges; (D) Each FCN operates on its specific feature maps and outputs classification (*cls*) and regression (*reg*) results; (E) The *cls* and *reg* outputs are converted into scored bounding boxes, and non-maximum suppression (NMS) is applied to all bounding boxes of all branches to obtain the final detection results.

In this paper, we aim to propose a face detector, whose design takes into consideration both computational efficiency and effectiveness. Regarding efficiency, we propose a multi-branch fully convolutional neural network architecture (MB-FCN), in which each branch is a FCN trained for detecting faces within specific scale ranges. And the MB-FCN detector can deal with faces at all scales in an image, while only requiring a single pass through the network. Compared to passing through each level of an image pyramid, our MB-FCN model can save computational resources and therefore is more efficient. Regarding effectiveness, we enable skip connections of the convolutional (*conv*) feature maps at different layers to represent faces in specific scale ranges. Specifically, small faces are represented with both shallow (spatially fine-grained) and deep (spatially coarse) features. In our experiments, we demonstrate that this representation is very important in achieving superior performance, especially on small faces.

**Contributions.** This paper makes three main contributions. **(1)** A new multi-branch FCN architecture for face detection is proposed, where each branch is a FCN trained to handle faces in specific ranges. More importantly, the MB-FCN detector can detect faces within all scale ranges in images with a single pass. **(2)** Scale specific skip connections of different layers are exploited for each branch, which can represent faces within specific ranges much better than only using the final *conv* feature maps, especially for faces at small scales. **(3)** The MB-FCN detector outperforms state-of-the-art methods on two popular benchmarks, where the most impressive improvement occurs in the most challenging subset of these datasets. Also, MB-FCN is computationally efficient with a runtime of 15 FPS on a GPU for $640 \times 480$ images and with no assumption on the minimum detectable face size.

## 2. Related Work

### 2.1. Handcrafted Feature Based Face Detection

As a classic topic, numerous face detection systems have been proposed during the past decade or so. Building an effective and efficient detector is the ultra goal of detection system designers. Since the seminal work of Viola and Jones (VJ) [27], the boosting cascade framework has become the *de facto* standard for rapid object detection. The attentional cascade structure is the critical component for the success of the VJ framework. The key insight is that smaller but more efficient classifiers can be learned, which can reject the majority of negative sub-windows, while keeping almost all positive examples. Consequently, most of the sub-windows will be rejected in early stages of the detector, making the system extremely efficient. However, due to the limited representation of Haar-like features, the original VJ detector shows poor performance in uncontrolled environments. HOG [5], SURF [17] and other sophisticated features [31] have been exploited to enrich the capacity of feature representation, and thus improve detection performance.

A similar idea has also been applied to cascaded deformable parts model (DPM) [30, 29], which is another traditional paradigm for object detection. In each stage of the cascade DPM detector, a hypothesis can be pruned if its score is below a pre-learned threshold. Therefore, this can reduce the number of parts evaluated and accelerate the evaluation process without sacrificing detection accuracy.

4322

However, most of the detection systems based on handcrafted features only train a single scale model, which is applied to each level of a feature pyramid, thus, increasing the computational cost drastically, especially for complicated features. More importantly, the limited representation of hand-crafted features restricts the performance of detectors, particularly in uncontrolled settings.

## 2.2. CNN-Based Face Detectors

In recent years and motivated by the superior performance of CNNs for image classification and scene recognition [15, 25, 39], generic object detectors based on CNNs, *e.g.*, the Region-based CNN (RCNN) [8], Faster RCNN [23] and its other variants have been introduced [8, 7, 23] and they achieve state-of-the-art detection performance. Specifically, Faster RCNN [23] has recently achieved a balance between both detection performance and computational efficiency. In fact, it has become the *de facto* framework for general object detection.

Inspired by the great success of Faster RCNN, several recent works [13, 28, 40] have utilized this framework to detect faces and shown impressive performance on the FDDB benchmark [11]. However, performance drops dramatically on the more challenging WIDER FACE dataset [33], which contains a large number of faces with lower resolution. The main reason for this disparity is that deep *conv* feature maps with lower spatial resolution are used for representation, which is insufficient for handling small faces [38, 2]. To overcome this problem, detectors [13, 28, 40] have to upsample input images during training and testing, which inevitably increases memory and computation costs. Compared to these methods, our detector exploits the intermediate *conv* feature maps with higher resolution to deal with small faces, making the process more efficient.

To deal with the efficiency problem, Li *et al.* [16] propose a cascaded architecture for real-world face detection inspired by the success of boosting-based algorithms [27]. Two lower-resolution models are used to quickly reject most hypothesis windows and higher resolution models are applied to get the final detection results. A multi-task variant of [16] for face detection and alignment is proposed in [37]. However, each stage in [16, 37] needs to be tuned carefully and is trained separately. To overcome this problem, a joint training variant is proposed in [22]. However, all these methods have an assumption on the minimum resolution of detected faces. Decreasing the minimum resolution quickly increases the runtime of these methods.

Compared to these methods, our detector learns multi-branch models to detect faces of all scales with a single pass through the backbone network. Hence, our detector is more effective. Unlike other methods, faces with lower spatial resolution (prevalent in the WIDER FACE dataset and in real-world applications) will not be missed by our detector, since it does not make any assumption on the minimum scale of faces to be detected.

## 2.3. Multi-Branch Generic Detectors

The works of [19, 2] employ intermediate *conv* feature maps with fine resolution to represent small objects and learn multi-scale models to detect small objects in a single shot. There are some common design aspects between their CNN architecture and ours. However, our method detects small objects by combining both shallow (spatially fine-grained) feature maps with deep and powerful (spatially low resolution) feature maps, which shows significant improvements over these two works [19, 2]. Moreover, we empirically show that adding more branches in [19, 2] might not increase the performance of face detection.

## 2.4. Skip Connection Representation

Small objects in images require fine scale representation. Skip connections [9, 1, 36] have been proposed for this purpose, which combines fine-grained *conv* feature maps from shallow layers and coarse semantic features from deep layers to represent objects more precisely. However, simply incorporating all feature maps from different layers may not yield performance improvement. In fact, increasing these skip connections might decrease detection performance in some cases.Unlike these previous methods, we make a thorough analysis of this design aspect to find the best skip connections for each branch in our face detection network.

## 3. Multi-Branch FCN Model

In this section, we will introduce our deep architecture for multi-branch FCN face detection and then give a detailed description on how to implement it.

### 3.1. Overview of Architecture

As illustrated in Figure 2, the whole architecture consists of four components. **(i)** The first component is the shared intermediate *conv* layers, which can be of any typical architecture like AlexNet [15], VGGNet [25] or ResNet [10]. In our experiments, ResNet-50 [10] pre-trained on the ImageNet dataset is adopted as the backbone network. After images are passed through the backbone network, the *conv* feature maps of every layer are generated. **(ii)** The second component creates the skip connections of feature maps at different scales to represent targets of different resolutions for every branch. Up-sampling and down-sampling operate on the *conv* feature maps of different scales to produce $K$-branch feature maps. Using this implicit multi-scale feature construction strategy can save a substantial amount of computational cost compared to explicitly re-sampling images, which have to pass through the backbone network several times. **(iii)** For each branch, multi-task learning is deployed to learn one FCN to deal with faces within certain scales. Specifically, each FCN takes as input a $1 \times 1$ spatial window of the input convolutional feature map, and outputs a lower-dimensional feature, which is fed into two sibling fully-connected layers (*reg* and *cls*) with filter size $1 \times 1$. **(iv)** The *reg* and *cls* outputs of every branch are converted to scored bounding boxes. Then, non-maximum suppression (NMS) is applied to those with confidence above a predefined threshold, and the final detection results are obtained.

### 3.2. Branch-Specific Skip Connection

In the forward pass, the backbone network computes a series of *conv* feature maps at several scales with a scaling step of 2. For clarity, we denote these feature maps as $\{C2, C3, C4, C5\}$ for conv2, conv3, conv4, and conv5 outputs of ResNet-50. The original stride of ResNet-50 is 32, which makes the final $C5$ feature maps too coarse. To increase the resolution of the $C5$ feature maps, we reduce the effective stride from 32 pixels to 16 pixels as done in [14]. Consequently, their corresponding strides of each layer are $\{4, 8, 16, 16\}$ pixels with respect to the input image.

In fact, $50\%$ of faces in the WIDER FACE dataset have an area less than $32 \times 32$ or even $16 \times 16$ pixels, so these objects will have been down-sampled to $2 \times 2$ or $1 \times 1$ at the $C5$ stage. Clearly, if this happens, most spatial information will have been lost due to the effective 16 time down-sampling. Therefore, skip connections are needed to guarantee the classifier access to information from features at multiple spatial resolutions. This will especially help detect small faces. To deal with different resolutions of different *conv* feature maps, deconvolution [20] using fixed bilinear interpolation weights and max pooling are used for up-sampling and down-sampling in our current implementation. We can also design other more complicated skip connection modules, however, these will inevitably sacrifice computational efficiency.

In our architecture, there are $K$ branches to deal with faces at different scales. Should we simply connect all the *conv* outputs $C2, C3, C4, C5$ to every branch? Obviously, it would be sub-optimal. First, it will increase the computational burden. Second, simply skipping connection of all intermediate *conv* maps might decrease the performance due to over-fitting caused by the curse of dimensionality. For example, the less powerful $C2$ feature maps might be useless for faces of large scales, thought its fine information might be essential in representing faces at lower resolution. Therefore, we need to find the best skip connections for each branch considering both efficiency and effectiveness. Interestingly, the architectures of MSCNN [2] and SSD [19] can be seen as a special version of our architecture, if there are no skip connections for each branch.

### 3.3. Multi-Branch Multi-Task Training

The $k$-th branch FCN has two sibling output layers, *cls* and *reg*. The *cls* layer computes the confidence score $y_{ki}$ of the $i$-th anchor at the $k$-th branch, where an anchor denotes a candidate box associated with a scale and aspect ratio [23]. The score $y_{ki}$ represents whether the corresponding anchor shows a face or not. Given the ground truth label $y^*_{ki} \in \{0, 1\}$, the softmax loss for classification is defined as follows:

$$L_{cls}(y_{ki}, y^*_{ki}) = y^*_{ki} \log(y_{ki}) + (1 - y^*_{ki}) \log(1 - y_{ki}) \tag{1}$$

The *reg* layer generates the 4 parameterized coordinates of the predicted bounding box $\mathbf{p}_{ki} = [p_x, p_y, p_w, p_h]_{ki}$, corresponding to its left corner, width, and height. Moreover, we represent the ground-truth box $\mathbf{p}^*_{ki} = [p^*_x, p^*_y, p^*_w, p^*_h]_{ki}$ associated with anchor $i$ of branch $k$. We utilize the standard regression loss proposed in [7], which is defined as follows:

$$L_{loc}(\mathbf{p}_{ki}, \mathbf{p}^*_{ki}) = \sum_{j \in \{x,y,w,h\}} \text{smooth}_{L_1}(\mathbf{p}^*_{ki} - \mathbf{p}_{ki})_j, \tag{2}$$

where

$$\text{smooth}_{L_1}(x) = \begin{cases} \frac{x^2}{2} & \text{if } |x| < 1 \\ |x| - 0.5 & \text{otherwise} \end{cases} \tag{3}$$

is a robust $L_1$ loss that is less sensitive to outliers than the $L_2$ loss. With these definitions, we can minimize the following multi-branch multi-task loss $L$ of anchors on the *conv* feature maps to jointly train for classification and bounding-box regression:

$$L(\mathbf{y}, \mathbf{p}) = \sum_{k=1}^{K} \gamma_k \sum_{i=1}^{N} (L_{cls}(y_{ki}, y^*_{ki}) + \lambda_k y^*_{ki} L_{loc}(\mathbf{p}_{ki}, \mathbf{p}^*_{ki})) \tag{4}$$

where $\mathbf{y} = [y_{1i}, ... y_{1N}, ..., y_{K1}, ..., y_{KN}]$ and $\mathbf{p} = [\mathbf{p}_{1i}, ... \mathbf{p}_{1N}, ..., \mathbf{p}_{K1}, ..., \mathbf{p}_{KN}]$ denote the vectors of predicted labels and bounding boxes of the corresponding anchors, respectively. $N$ denotes the mini-batch size. $\gamma_k$ balances the importance of models at different branches. In our experiments, each $\gamma_k$ is set to 1, which means that all $K$ models carry the same importance. This can be changed to reflect the scale statistics of faces in the training data. The term $y^*_{ki} L_{loc}(\mathbf{p}_{ki}, \mathbf{p}^*_{ki})$ means that the regression loss is activated only for positive anchors (*i.e.* when $y^*_{ki} = 1$) and is disabled otherwise. For background anchors, there is no notion of a ground-truth bounding box and hence $L_{loc}$ is ignored. $\lambda_k$ is the tradeoff parameter between classification and localization. In our experiments, we set to $\lambda_k = 2 \; \forall k$ to encourage better localization.

### 3.4. Implementation Details

**Training setup.** We use the large-scale WIDER FACE dataset [33] to train our MB-FCN detector. During training, each mini-batch is constructed from one image, chosen uniformly at random from the training dataset. To fit it in GPU memory, the image is resized by the ratio $1024/\max(w,h)$, where $w$ and $h$ are its width and height, respectively.

An anchor is assigned with a positive label, if the intersection over union (IoU) overlap between it and any ground-truth bounding box is larger than $0.55$; otherwise, it is negative if the maximum IoU with any ground-truth face is less than $0.35$. We also employ data augmentation by horizontally flipping each image per batch with a probability of $0.5$.

**Optimization.** Our MB-FCN model is a fully convolutional network [20], which can be trained end-to-end by back-propagation and stochastic gradient descent (SGD). We follow the image-centric training strategy from [7, 23] in training. Each mini-batch contains many positive and negative examples that are sampled from a single image. Obviously, negative samples will dominate, which will lead to biased prediction, if they are all used to compute the loss function. To avoid this bias and to speedup training, we employ the hard negative example mining strategy in [14, 24], where different sets of negative samples are chosen in each iteration based on their loss using the most recently trained network.

**Hyper-parameters.** The weights of the filters of all layers (except for the shared ones) are initialized by randomly drawing from a zero-mean Gaussian distribution with standard deviation $0.01$. Biases are initialized at $0.1$. All other layers are initialized using a model pre-trained on ImageNet. The same mini-batch size of $128$ is employed for each branch in MB-FCN. The learning rate is initially set to $0.001$ and then reduced by a factor of $10$ after every $30k$ mini-batches. Training is terminated after a maximum of $80k$ iterations. We also use a momentum of $0.9$ and a weight decay of $0.0005$. Our system is implemented in Caffe [12] and its source code will be made publicly available.

## 4. Experiments

In this section, we will experimentally validate our proposed method. Firstly, we dive into the details to find the best skip connections for each branch and the optimal number of branches for our detector. Secondly, we evaluate the proposed MB-FCN detector on two public face detection benchmarks, including WIDER FACE [33] and FDDB [11], while comparing it against state-of-the-art detectors.

### 4.1. Training and Validation Datasets

To train our multi-branch FCN detector, we use a recently released large-scale face detection benchmark, the WIDER FACE dataset [33]. It contains $32,203$ images, which are selected from the publicly available WIDER dataset. And $40\%/10\%/50\%$ of the data is randomly selected for training, validation, and testing, respectively. There are $393,703$ labeled faces with a high degree of variability in scale, pose, occlusion, expression, appearance, and illumination. Images in WIDER FACE are categorized into 61 social event classes, which have much more diversity and is closer to what is encountered in real-world applications. Therefore, we use this dataset for training and validating our models under different parameter settings.

In fact, the WIDER FACE dataset is divided into three subsets: Easy, Medium, and Hard, based on the heights of the ground truth faces [33]. The Easy/Medium/Hard subsets contain faces with heights larger than 50/30/10 pixels respectively. Compared to the Medium subset, the Hard one contains many faces with height between $10-30$ pixels. As expected, it is quite challenging to achieve good detection performance on the Hard subset.

### 4.2. What are the best skip connections for each branch?

Skip connections combine the intermediate *conv* feature maps for precise representation of objects, especially those with small spatial resolution. However, connecting all these maps together will likely make the detector inefficient and the avoidance of over-fitting challenging. As such, we seek the optimal connections for each branch. To do this, we first seek the best combination of connections for each single branch separately. Then, we perform a greedy strategy that loops over each branch optimizing its best connections while keeping the other connections the same. In doing so, we are iteratively improving the overall detection performance while greedily adding connections to the different branches. First, we summarize the detection results for the single branch models in Table 1, in which case they are applied independently. It is clear that each of these models achieves impressive validation performance on all subsets of WIDER FACE dataset [33], especially when compared to three state-of-the-art face detectors [33, 37, 40]. In fact, the C5(16) model, which only uses the feature maps from the final $C5$ layer, nearly surpasses all other methods on all subsets. It only shows slightly worse results on the Hard subset as compared to the CMS-RCNN detector [40], which combines three *conv* feature maps of VGGNet [25].

Compared to the C5(16) model, adding the $C4$ feature maps slightly increases the performance on all subsets (refer to the $4^{th}$ and $5^{th}$ rows of Table 1). Moreover, connecting layers closer to the bottom of the network ($C2$ and $C3$) hardly produces any performance gain, except on the Hard subset. We can conclude that one branch models are designed specifically for the Easy and Medium subsets, but more effort needs to be made to improve their performance on the Hard subset. Therefore, in what follows, we use the skip connections of $C4$ and $C5$ as the optimal settings for single branch models.

Table 1. Performance of our single branch models with different skip connections on the WIDER FACE validation set broken down into three subsets. The average precision (AP) results are reported. In the model CX(Y), X denotes which layer(s) is (are) used for skip connection and Y stands for the stride of the feature maps. The same annotation is used in Table 2 & 3.

| Skip Connection | Easy | Medium | Hard |
|---|---|---|---|
| Multi-scale CNN [33] | 0.711 | 0.636 | 0.400 |
| Multi-task CNN [37] | 0.851 | 0.820 | 0.607 |
| CMS-RCNN [40] | 0.902 | 0.874 | 0.643 |
| C5(16) | 0.920 | 0.887 | 0.640 |
| C45(16) | 0.921 | 0.888 | 0.643 |
| C345(16) | 0.921 | 0.888 | 0.645 |
| C2345(16) | 0.921 | 0.886 | 0.645 |

Table 2. AP results of our two branch models with different skip connections on the WIDER FACE validation set.

| Skip Connection | Easy | Medium | Hard |
|---|---|---|---|
| C3(8)-C45(16) | 0.911 | 0.888 | 0.735 |
| C35(8)-C45(16) | 0.917 | 0.905 | 0.782 |
| C345(8)-C45(16) | 0.918 | 0.907 | 0.783 |
| C2345(8)-C45(16) | 0.917 | 0.907 | 0.782 |

Table 3. Performance of our different branch models with different skip connection on the WIDER FACE validation set. The AP results are reported.

| Skip Connection | Easy | Medium | Hard |
|---|---|---|---|
| C45(16) | 0.921 | 0.888 | 0.643 |
| C345(8)-C45(16) | 0.918 | 0.907 | 0.783 |
| C345(8)-C45(16)-C45(32) | 0.916 | 0.905 | 0.774 |

Although single branch models achieve surprising performance on the Easy and Medium subsets, the results on the Hard subset are far from satisfactory. Hence, we add another branch to detect faces with smaller spatial size. We summarize the detection results of these two branch models in Table 2. In this table, we observe that the C3(8)-C45(16) model (*i.e.* a model with one branch connected to $C4$ and $C5$, while the other branch connected to $C3$) shows significant improvement (more than 9% in absolute AP) over the C45(16) model on the Hard subset. This indicates that feature maps with finer spatial resolution are helpful in detecting faces at lower resolution and that the added branch enables this improvement. However, for this comparison, the performance of the two branch model drops on the Easy subset. After checking the detection results, we find that this decrease is mainly caused by false positive results with high classification score, which are generated from the new branch (*i.e.* small false positive faces). The C3(8)-C45(16) model utilizes the less powerful $C3$ feature maps to represent the small objects, which is similar in spirit to SSD [19] and MSCNN [2]. When we up-sample the deep feature map $C5$ and connect it with the $C3$ features (leading to the C35(8)-C45(16) model), the performance on all three subsets increases, especially on the Medium and Hard subsets (the $1^{st}$ and $2^{nd}$ rows of Table 2). This demonstrates that the connections of both shallow and deep feature maps are powerful and representative for small faces. When the $C4$ feature maps are added, the performance improves only slightly (the $3^{rd}$ and $2^{nd}$ rows of Table 2). Moreover, adding the $C2$ feature maps does not improve performance at all.

From the above comparisons, we conclude that skip connections can increase the performance on faces at all scales. However, we should design the connections carefully for each branch rather than simply connecting *conv* feature maps of all layers. Therefore, we conclude that combining deep (and spatially coarse) feature maps with shallow (and spatially fine-grained) feature maps is the key to successfully detecting faces at small scales.

### 4.3. What is the optimal number of branches?

In [2], the authors claim that the inconsistency between the sizes of objects and receptive fields compromises detection performance. Therefore, they add extra layers with large receptive fields to detect objects at large scale. However, there is no gain in performance when we add extra branches, as shown in Table 3. The reason might be that the deep $C5$ feature maps are of large receptive fields and are already representative for face detection, as compared to generic object detection [19, 2]. In fact, detection performance degrades, when a third branch is added (refer to last row of Table 3). This is most probably due to issues in training such a large network. Finding a favorable setting for all the hyper-parameters in this case is challenging. Therefore, we argue that two branches are enough for face detection and, unlike SSD [19] and MS-CNN [2], no extra branches are needed. So, in what follows, our MB-FCN detector is chosen to be the two branch C345(8)-C45(16) model.

### 4.4. Evaluation on FDDB [11]

The FDDB dataset [11] is a challenging benchmark for face detection. And it is arguably the most popular benchmark for face detection and nearly all state-of-art face detectors have been tested on it. FDDB only uses precision at specific false positive rates rather than AP (average precision) to evaluate detectors, which is not necessarily the most comprehensive way of evaluating detection performance. However, we follow convention and use this metric to compare with other methods. There are 67 unlabeled faces in FDDB [3], making precision not accurate at small false positive rates (*e.g.* @100fp). Hence, we report the precision rate at 500 false positives. Our MB-FCN detector (*i.e.* the C345(8)-C45(16) two branch model) achieves a superior performance over all other CNN-based detectors.

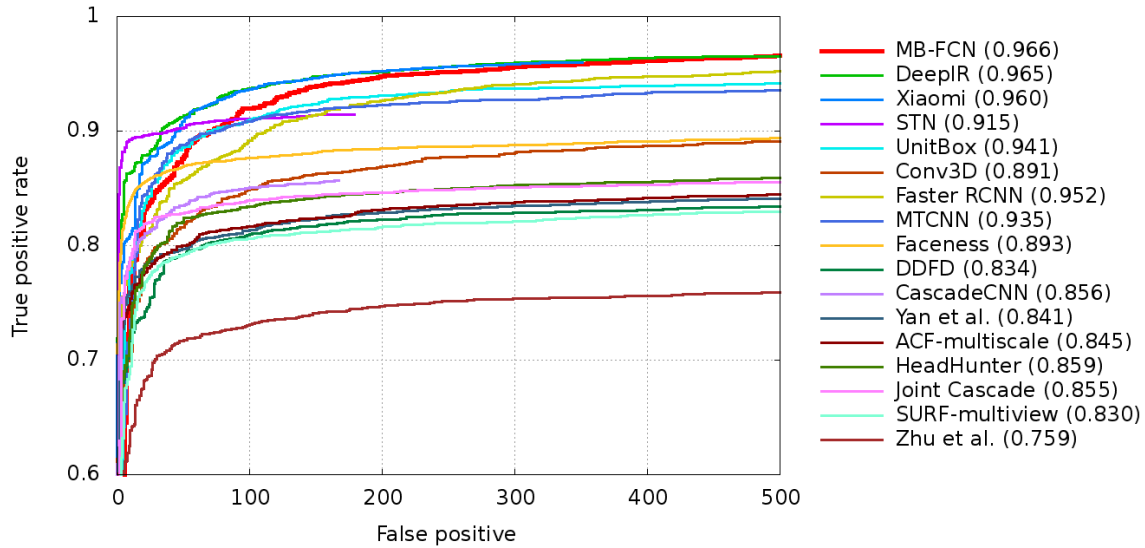

Figure 3. On the FDDB dataset, we compare our MB-FCN detector against many state-of-the-art methods: DeepIR [26], Xiaomi [28], STN [3], UnitBox [34], Conv3D [18], Faster RCNN [13], MTCNN [37], Faceness [32], DDFD [6], CascadeCNN [16], Yan *et al.* [29], ACF-multiscale [31], HeadHunter [21], Joint Cascade [4], SURF-multiview [17] and Zhu *et al.* [41]. The precision rate with 500 false positives is reported in the legend. The figure is best viewed in color.

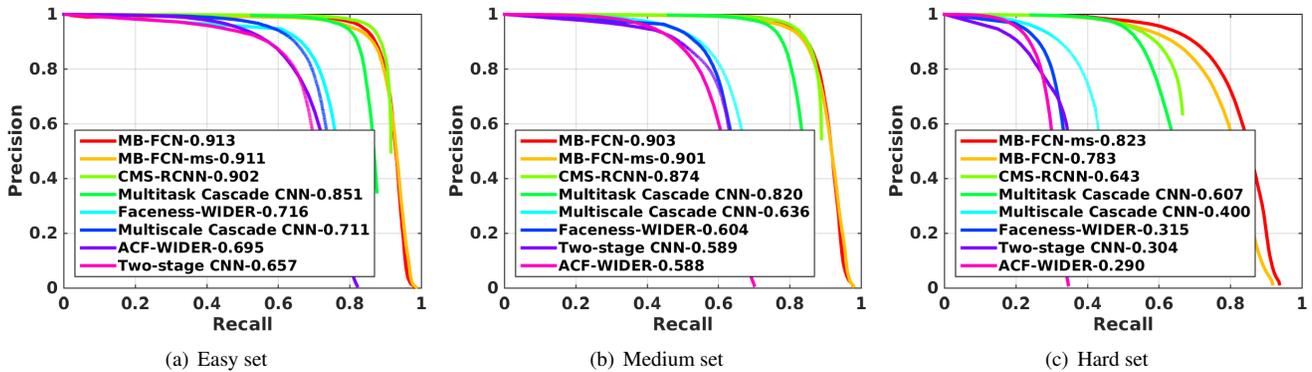

(a) Easy set  (b) Medium set  (c) Hard set

Figure 4. On the WIDER FACE dataset, we compare our MB-FCN detector against several state-of-the-art methods: CMS-RCNN [40], Multi-task Cascade CNN [37], Faceness-WIDER [32], Multi-Scale Cascade CNN [33], Two-Stage CNN [33], and ACF-WIDER [31]. The average precision (AP) results are reported in the legend. The figure is best viewed in color.

These results are shown in Figure 3. Moreover, the MB-FCN detector even surpasses all other detectors in precision at small false positive rates (*e.g.* @100-400fp) except the Xiaomi [28] and DeepIR [26] detectors. However, both of them are trained on WIDER FACE and fine-tuned on FDDB, while MB-FCN is only trained on WIDER FACE.

### 4.5. Evaluation on WIDER FACE [33]

Now, we apply our MB-FCN detector to the testing set of WIDER FACE. The results of the Easy/Medium/Hard testing subsets are shown in Figure 4. We see that the proposed MB-FCN detector surpasses all the methods on all subsets especially for the Hard subset, which is by far the most challenging. In comparison, CMS-RCNN [40] is a variant of Faster RCNN [13], which utilizes multi-layer *conv* feature maps and context information to detect faces with low resolution. Compared to CMS-RCNN, MB-FCN achieves slightly better (+1.2%, +2.9%), and much better performance (+14.0%) on the Easy, Medium, and Hard subsets respectively. In fact, the improvement is quite significant on the latter subset, which further motivates our multi-branch approach capable of accurate face detection at small scales. We also test our MB-FCN detector on a coarse-level image pyramid (*i.e.* the input image is sampled at different scales and pushed through MB-FCN several times), which is a strategy used by previous face detectors. We denote this strategy as MB-FCN-ms. It is not surprising to see that it achieves better (+4.0%) performance than MB-FCN on the Hard subset, while running about five times slower.

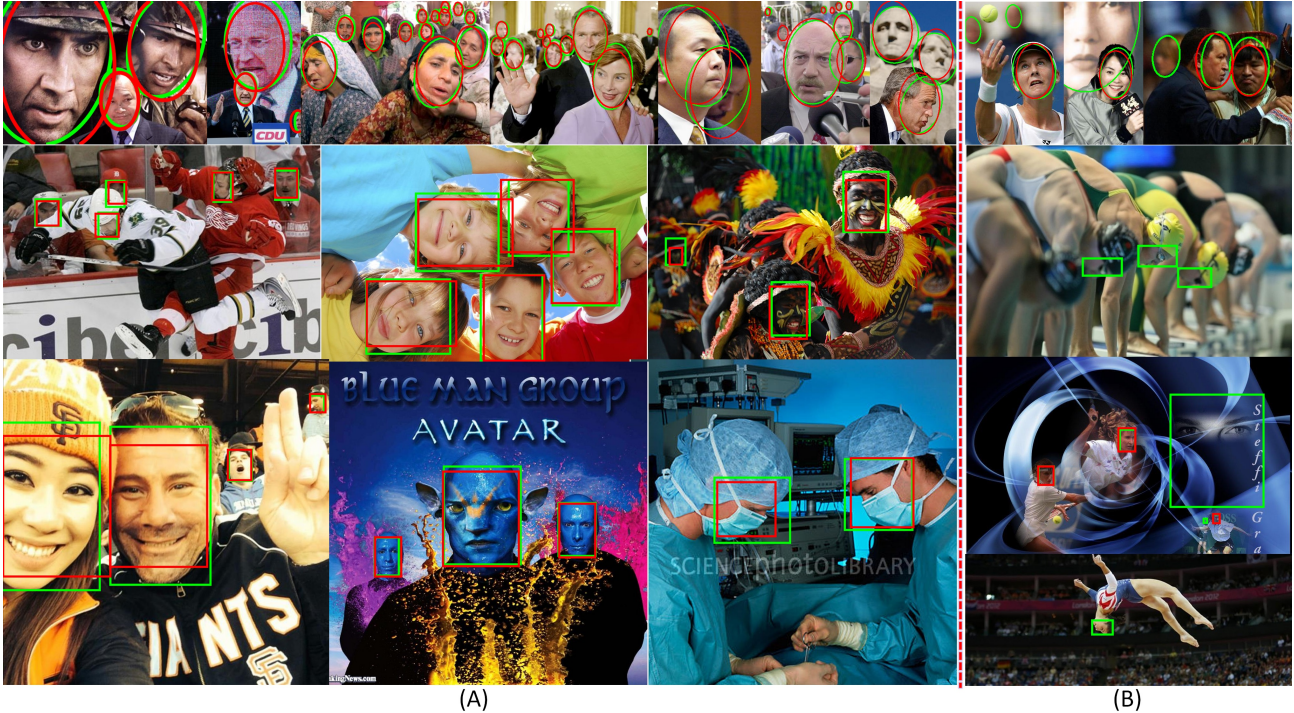

Figure 5. Some example results of the proposed MB-FCN face detector. Bounding boxes with green or red color are ground-truth annotations and MB-FCN detection results, respectively. (A) From the successful cases, we see that MB-FCN can deal with faces with extreme poses, large scale variations, exaggerated expressions, severe makeup and occlusion; (B) Some faces with extreme pose, severe truncation, or low resolution can still cause failures for MB-FCN.

### 4.6. Runtime Evaluation

One of the important advantages of our MB-FCN detector is its efficiency in dealing with faces in a wide range of possible scales. During the detection process, an image passes through the backbone network in only one single shot, which is more efficient than other methods that use an image pyramid. CascadeCNN [16, 37] is also designed with efficiency in mind and it runs at 100 FPS on a GPU for VGA (640×480 pixels) images. However, this speed is reported based on the assumption that it only encounters faces with resolution higher than $80 \times 80$ pixels. With this assumption, many faces with lower resolution would be missed, which is especially the case in the WIDER FACE dataset and in real-world applications. Decreasing the minimum detectable face size in CascadeCNN quickly increases the runtime of this method[1]. Currently, our detector runs at 15 FPS on VGA images with no assumption on the minimum detectable face size. Moreover, the fast version of the STN detector [3] runs at about 30 FPS on VGA images. However, it only handles faces larger than $36 \times 36$ pixels and the ROI convolution is used to speed up the detector with minimal impact on the recall. As such, ROI convolution can also be employed to further accelerate our detector.

### 4.7. Qualitative Results

We show some qualitative face detection results on sample images in Figure 5. From Figure 5(A), we observe that our MB-FCN detector can deal with challenging cases that have extreme poses, large scale variations, exaggerated expressions, severe makeup, and occlusion. However, Figure 5(B) also shows some failure cases, which are caused by very challenging nuisances. These results indicate that more progress is needed to further improve face detection performance, both in accuracy as well as runtime.

## 5. Conclusion

In this paper, we propose a multi-branch fully convolutional network (MB-FCN) for face detection. In our detection system, each branch uses specific connections of different *conv* layers to represent faces and learn a separate branch FCN for specific scale ranges. More importantly, the MB-FCN detector can detect faces within all scale ranges in a single shot, which makes it computationally efficient as it runs at 15 FPS on VGA images. Our MB-FCN detector is evaluated on two public face detection benchmarks, including FDDB and WIDER FACE and achieves superior performance when compared to state-of-the-art face detectors.

---

[1] On a workstation with Intel CPU E5-2698 and NVIDIA TITAN X GPU, CascadeCNN runs at 46, 20, and 10 FPS at a minimum detectable face size of $80 \times 80$, $20 \times 20$ and $10 \times 10$ pixels, respectively.